\documentclass[preprints,article,accept,pdftex,moreauthors]{Definitions/mdpi} 

\newcommand\blfootnote[1]{%
 \begingroup
 \renewcommand\thefootnote{}\footnote{#1}%
 \addtocounter{footnote}{-1}%
 \endgroup
}
\firstpage{1} 
\pubvolume{28}
\issuenum{8}
\articlenumber{925}
\pubyear{2026}
\copyrightyear{2026}
\externaleditor{} % More than 1 editor, please add `` and '' before the last editor name
\datereceived{3 July 2026} 
\daterevised{11 August 2026} % Comment out if no revised date
\dateaccepted{15 August 2026} 
\datepublished{18 August 2026 } 

\Title{Comparison of D-Wave Quantum Annealing and Gibbs Monte Carlo for Sampling from a Probability Distribution of a Restricted Boltzmann Machine}

\Author{Abdelmoula El-Yazizi
 *\orcidA{} and Yaroslav Koshka}

\AuthorNames{Abdelmoula El-Yazizi and Yaroslav Koshka}

\address[1]{%
Department of Electrical and Computer Engineering, Mississippi State University, Starkville, MS 39762, USA; ykoshka@ece.msstate.edu
}
\corres{\hangafter=1 \hangindent=1.05em \hspace{-0.82em}Correspondence: aelyazizi@gmail.com}

\abstract{A local-valley (LV)-centered approach to assessing the quality of sampling from Restricted Boltzmann Machines (RBMs) was applied to the latest generation of the D-Wave quantum annealer. D-Wave and Gibbs samples from a classically trained RBM were obtained at conditions relevant to contrastive-divergence-based RBM learning. The samples were compared for the number of  LVs to which they belonged and the energy of the corresponding local minima. No significant (desirable) increase in the number of the LVs has been achieved by decreasing the D-Wave annealing time. At any training epoch, the states sampled by the D-Wave belonged to a somewhat higher number of LVs than in the Gibbs sampling. However, many of those LVs found by the two techniques differed. For high-probability sampled states, the two techniques were (unfavorably) less complementary and more overlapping. Nevertheless, many potentially “important” local minima, i.e., those having intermediate, even if not high, probability values, were found by only one of the two sampling techniques while missed by the other.  The two techniques overlapped less at later than earlier training epochs, which is precisely the stage of the training when modest improvements to the sampling quality could make meaningful differences for the RBM trainability. The results of this work may explain the failure of previous investigations to achieve substantial (or any) improvement when using D-Wave-based sampling. However, the results reveal some potential for improvement, e.g., using a combined classical--quantum approach.}


\usepackage{braket}
\usepackage{longtable}

\begin{document}

\blfootnote{Published in \textit{Entropy} 28 (2026), 925. doi:10.3390/e28080925}
\section{Introduction} \label{sec1}
Recently, substantial efforts have been invested by the machine learning (ML) community in using quantum annealers (QAs), which primarily included the first-in-the-world commercial quantum computer (QC) by  D-Wave Inc. ( Palo Alto, CA, USA)
, for sampling from complex probability distributions, including those of Boltzmann Machines (BMs), Deep BMs (DBMs) and other probabilistic graphical models \cite{adachi_application_2015,amin_quantum_2018,dixit_training_2021,dixit_training_2022,dorband_boltzmann_2015,korenkevych_benchmarking_2016,benedetti_quantum-assisted_2017,caldeira_restricted_2020,perdomo-ortiz_opportunities_2018,sleeman_hybrid_2020,rocutto_quantum_2021,dumoulin_challenges_2014}. For example, in previous work by the authors’ group \citep{dixit_training_2022}, training of Restricted Boltzmann Machines (RBMs) using a QA from D-Wave Inc. was conducted using an approach proposed by Adachi et al.
 \cite{adachi_application_2015}. In that approach, an assumption was that “samples” produced by the D-Wave follow a Boltzmann probability distribution, albeit at a temperature (\textit{T}) much lower than $T=1$ used in ML. This called for estimating and adjusting the sampling temperature following the algorithm proposed in \cite{adachi_application_2015}. When the D-Wave was used for sampling from the RBM’s probability distribution during gradient descent optimization of the log-likelihood, a substantial improvement in DBM training was achieved \cite{adachi_application_2015}. However, many other BM training experiments showed either comparable to the classical \cite{benedetti_estimation_2016}, or somewhat improved \cite{korenkevych_benchmarking_2016,rocutto_quantum_2021}, or somewhat inferior \cite{dixit_training_2021}  training performance.

Previous work by the authors' group aimed at understanding both the opportunities and shortcomings of a QA-based sampling, including investigating the differences in the sampling “quality” when using the D-Wave QA versus the classical Gibbs sampling based on Markov Chain Monte Carlo (MCMC). When one refers to a “D-Wave sample”, it usually means a set of solution-repetitions returned by the D-Wave, which can be used either as a sample itself (at some unknown sampling temperature $T$) \cite{adachi_application_2015} or as a set of seeds for classical Monte Carlo (MC)  \cite{korenkevych_benchmarking_2016,kalis_hybrid_2023}. While there have been reports of attempts to compare D-Wave and MC samples using rigorous statistical measures (e.g., Kullback--Leibler (KL) divergence \cite{goto_online_2023} or the expectation values of selected observables that are diagonal in the computational basis \cite{gonzalez_testing_2021}), a different approach proposed and explored by the authors of this work
 \cite{koshka_comparison_2020,koshka_toward_2020,koshka_comparison_2021} aimed at investigating differences between the D-Wave and Gibbs samples that do not rely upon traditional quantitative probabilistic measures but was found to be particularly relevant when applied to undirected probabilistic graphical models (and BMs in particular). For BMs (including RBMs), many local valleys (LVs) are created in the multidimensional configuration space during training. Some of those are LVs with local minima (LMs) that (by learning) emerge near the RBM Training Patterns (TPs). A sample that consistently misses many of those LVs (especially when they have low energy/high probability) could cause incorrect classification of similar patterns located inside the particular LV. Moreover, there are “spurious” LVs containing undesirable high-probability (low-energy) states that also must be “visited” by the sampling to allow the training algorithm to either eliminate them or at least reduce the associated probabilities. Therefore, the LV-centered approach of this work includes analyzing the ability of the D-Wave’s sample to reveal potentially difficult-to-find low-energy/high-probability LVs. It should be noted that our focus on the statistics of LVs is closely related to the broader framework of Energy Landscape Theory, which has been extensively developed in statistical physics and chemistry and has also been applied to machine learning to characterize the optimization and generalization properties of neural networks \cite{ballard_energy_2017,choromanska_loss_2015} (see Section \ref{sec2.1} for additional discussion).

Two main shortcomings of our previous work \cite{koshka_comparison_2020,koshka_toward_2020,koshka_comparison_2021}  motivated the continuation of the LV-centered investigation of the sampling potential of the D-Wave compared to Gibbs sampling. First, the most rigorous comparison of the two sampling techniques was done when a set of findable LVs was obtained after a very prolonged exhaustive search utilizing a wide range of sampling temperatures ($T$) to find as many LVs as possible, both low- and high-energy~\cite{koshka_comparison_2020}. This is very different from what is going on during a \textit{k}G-step Contrastive Divergence RBM learning (CD-\textit{k}) \cite{hinton_training_2002,fischer_training_2014}. For practical RBM training applications and also for understanding why, in many cases, the D-Wave-based sampling has failed to deliver significant or even any improvements, it is crucial to conduct this comparison for identical-size samples and for the specific kind of classical sampling that is used during CD-\textit{k}, with \textit{k}G Gibbs sampling steps at $T=1$.

Second, all the experiments in our previous works~\cite{koshka_comparison_2020,koshka_toward_2020,koshka_comparison_2021}  were conducted using the previous generation of the D-Wave hardware employing a Chimera graph, a smaller number of qubits, and minimal connectivity between qubits. This hardware allowed us to embed only RBMs having impractical, small graph sizes. The new generation of the D-Wave hardware (the Pegasus graph) has increased the number of qubits (5760
 compared to 2041 in the D-Wave version used in our previous works~\cite{koshka_comparison_2020,koshka_toward_2020,koshka_comparison_2021}) and, especially, the connectivity (15 connections for a single qubit on average, compared to 6 in the Chimera D-Wave). This allowed us to investigate much larger RBMs compared to our previous works \cite{koshka_comparison_2020,koshka_comparison_2021}. The results in this work, however, are reported for a larger RBM, with 74 visible units ($N_v = 74$) and 74 hidden units ($N_h = 74$) (for $8\times8$-pixel images), as well as $N_v = 154$ and $N_h = 72$ (for $12\times12$-pixel images) in some of the experiments, compared to 64 for both $N_v$ and $N_h$ in our previous work. No less importantly, there have been studies~\cite{pelofske_comparing_2023} of a continuous improvement of the ability of the next generations of the QAs to find the Ground State (GS), which is the lowest-energy/highest-probability state of the Hamiltonian (i.e., of the cost function). While improving the GS search is the most welcome news concerning the primary purpose of a QA (i.e., optimization tasks), this improvement could mean unwelcome news for those applications where the QA is used for sampling. If an increased percentage of the D-Wave solution-repetitions return the same state, e.g., the GS, the entire sample offers much-decreased variance, and potentially a much lower representation of high- and moderate-probability states that a good sample of a given size should reveal. 

The primary motivation of this work was to address the two shortcomings of our previous works mentioned above concerning the need of RBM training: (1) compare the D-Wave and Gibbs samples from a classically trained (i.e., trained without the help from the QA) RBM at conditions relevant to sampling during the CD-\textit{k} RBM learning and (2) do it with the latest version of the D-Wave hardware and a larger RBM graph. Further, additional results will be presented on manipulating the QA parameters (the annealing time) to increase the sample variance. For the approach used in this work, increasing the sample variance primarily means trying to increase the number of findable LVs ($N_{\text{LV}}$) to which RBM states sampled by the D-Wave belong. As an illustration, while many LMs may exist, a sample (in this case, a D-Wave sample) could return only states that belong to a small number of LVs, even if all have low energy (high probability). In that case, even if all those sampled states followed a Boltzmann probability distribution, the quality of such a sample would be far from sufficient to use it in RBM learning successfully. 

The remainder of the paper is organized as follows. Section~\ref{sec2} introduces the necessary background information about the study. In  Section~\ref{sec3}, we will cover the methodologies utilized for the study, while Section~\ref{sec4} provides a thorough comparative analysis and discussion, which is followed by concluding the paper (Section~\ref{sec5}). 
\section{Background}\label{sec2}
\subsection{Local Valley Formation During BM Training}\label{sec2.1}

This section justifies our interest in LVs when comparing the D-Wave-based and the classical sampling. The joint probability distribution in the RBM model is the Gibbs distribution: 

\begin{equation}
  p(v,h) = \frac{1}{Z} e^{-\frac{E(v,h)}{T}},
\label{eq1}
\end{equation}

\noindent where $v$, $h$ are the vectors of the visible and hidden units, respectively. $E(v, h)$ is the RBM’s energy function, $Z(T)$ is the partition function, and $T$ is a temperature-like parameter. The RBM’s energy function is a particular case of a general Ising spin glass model's energy function, making the D-Wave a very convenient tool for RBM-related tasks. It has the form: 

\begin{equation}
E(v,h) = - \sum_{i=1}^{n} \sum_{j=1}^{m} \omega_{ij} h_i v_j 
- \sum_{j=1}^{m} b_j v_j 
- \sum_{i=1}^{n} c_i h_i,
\label{eq2}
\end{equation}	

\noindent where $v_{j}$ is a visible unit, $h_{i}$ is a hidden unit, $b_{j}$  and $c_{i}$ are the biases corresponding to the visible unit $v_{j}$ and hidden unit $h_{i}$, respectively.  $\omega_{ij}$ is the weight between visible unit $j$ and hidden unit $i$.
The partition function is 

\begin{equation}
Z(T) = \sum_{v,h} e^{-\frac{E(v,h)}{T}},
\label{eq3}
\end{equation}

The RBM training, as well as training of any other energy-based model, aims at maximizing the probability for those states that correspond to the TPs. This is accomplished by minimizing a loss function and quantifying a difference between the model and the unknown probability distributions describing the TPs. The marginal probability distribution $p(v)$ is maximized for the TPs $v_{\text{tr}}$ while keeping it low for all the other possible configurations. Concerning the focus on LVs adopted in this work, it is relevant to mention that the maximum $p(v)$ at $v = v_{\text{tr}}$ (i.e., the goal of the training) could, in general, be achieved while also having the joint probability $p(v,h)$ not having its maximum values at $v = v_{\text{tr}}$. This means that the LM of $E(v,h)$ would not necessarily have $v$ equal to one of the TPs. However, such an outcome during the training by log-likelihood maximization would be counteracted by the trend of the entropy maximization \cite{jaynes_information_1957}. Therefore, the probability of achieving the GS and LMs at exactly the states corresponding to TPs must increase during the training. It is to be noted that the maximum entropy that would still be consistent with the training goals may not be large enough to guarantee such an outcome, and LMs could have $v \not= v_{\text{tr}}$. Nevertheless, as was also confirmed in our previous work \cite{koshka_determination_2017}, most of the LMs during the RBM training often form at or at least in the vicinity of a TP. ~Some LVs, of course, also incorporate states corresponding to multiple TPs.

The following is a more detailed discussion of this process. Maximization of $p(v,h)$ is accomplished by the gradient ascent-based optimization of the model parameters $\theta$ to maximize the log-likelihood given the training data $\ln L(\theta \mid v_{\text{tr}}) = \ln p(v_{\text{tr}} \mid \theta)$. For an RBM, the parameters $\theta$ are $\omega_{ij}$, $b_{j}$,  and $c_{i}$. 
The corresponding expressions for the gradient of the log-likelihood for a single TP $v_{\text{tr}}$ are 

\begin{equation}
\frac{\partial \ln L(\theta | v_{\text{tr}})}{\partial \omega_{ij}} = p(H_i = 1 | v_{\text{tr}}) v_j - \sum_{v} p(v) p(H_i = 1 | v) v_j,
\label{eq4}
\end{equation}

\begin{equation}
\frac{\partial \ln L(\theta | v_{\text{tr}})}{\partial b_j} = v_j - \sum_{v} p(v) v_j,
\label{eq5}
\end{equation}

\begin{equation}
\frac{\partial \ln L(\theta | v_{\text{tr}})}{\partial c_i} = p(H_i = 1 | v_{\text{tr}}) - \sum_{v} p(v) p(H_i = 1 | v),
\label{eq6}
\end{equation}

The most computationally intensive parts of Equations~(\ref{eq4})--(\ref{eq6}) are the second terms, which generally require summation over all possible values of the vector of the visible units. While QA-based algorithms promise efficient computation for this part of the RBM learning algorithm, it is beyond the scope of this work. Instead, the D-Wave machine was applied in this work to perform QA on the Ising spin glass problem of an already trained RBM. For RBM training in this work, a standard approximation technique, Gibbs Sampling, was used (Section~\ref{sec3.1}), which is one of the MCMC algorithm implementations. This algorithm uses a TP as the initial value for the visible units in the Gibbs chain. The following expressions give the conditional probability of a single variable being equal to one. The Equations (\ref{eq7})--(\ref{eq9}) are used to sample sequentially at $T=1$ the next value of the vector of the visible or hidden units in the Gibbs sampling until either the stationary distribution or a specified limit of steps was reached:
\vspace{-6pt}
\begin{equation}
p(H_i = 1 | v) = \sigma \left( \sum_{j=1}^{m} \omega_{ij} v_j + c_i \right),
\label{eq7}
\end{equation}

\begin{equation}
p(V_j = 1 \mid h) = \sigma \left( \sum_{i=1}^{n} \omega_{ij} h_i + b_j \right),
\label{eq8}
\end{equation}

\begin{equation}
\sigma(x) = \frac{1}{1 + e^{-\frac{x}{T}}},
\label{eq9}
\end{equation}

\noindent where $\sigma(x)$ is the sigmoid function. 

When simulated annealing (SA) is applied, the same equations are used to find the states corresponding to the global and local minima. 
When maximizing the log-likelihood $\ln L(\theta \mid v_{\text{tr}}) = \ln p(v_{\text{tr}} \mid \theta)$ given the training data, the gradient ascent algorithm aims at maximizing the sum over all hidden units of joint probabilities corresponding to the TPs, which for a single training vector $v_{\text{tr}}$ has the following form:

\begin{equation}
p(v_{\text{tr}} \mid \theta) = \sum_{h} p(v_{\text{tr}}, h \mid \theta) = \frac{1}{Z(\theta)} \sum_{h} e^{-E(v_{\text{tr}}, h \mid \theta)},
\label{eq10}
\end{equation}

As a result, such training favors those changes that may maximize individual joint probabilities $p(v_{\text{tr}}, h \mid \theta)$ in the sum (i.e., minimize-$E(p(v_{\text{tr}}, h \mid \theta))$ at the expense of other joint probabilities corresponding to states $(v, h)$ other than the TPs. While it is far from guaranteed, a significant degree of $L(\theta \mid v_{\text{tr}})$ maximization makes it more likely that an LM of $E(v, h)$ will turn out to be a state $(v, h)$ corresponding to one of the TPs $v_{\text{tr}}$, while possibly including other (e.g., similar) TPs \cite{koshka_determination_2017}. This means that when approaching the end of the training, higher-probability states (i.e., states sampled most frequently from the RBM probability distribution by an efficient sampler) have visible vectors close to most of the TPs. This makes it efficient to follow the sampling approach used in the CD-\textit{k} training, which is to start the Gibbs sampling from each TPs. At the initial stages of the training, however, this outcome may still be far from reality, possibly making alternative sampling approaches more promising. 

As was demonstrated in~\cite{koshka_comparison_2020}, there is a substantial variation in the width of the LVs formed because of the RBM training. This means there is a considerable variation in the density of states at or close to the bottoms of the LVs, including those critical LVs incorporating LM-states with low energy (i.e., high probability). The concept of LVs is closely related to the broader framework of Energy Landscape Theory, which has been extensively developed in statistical physics and chemistry and has also been applied to machine learning to characterize the optimization and generalization properties of neural networks \cite{ballard_energy_2017,choromanska_loss_2015}. Within this framework, the objective of a sampling algorithm is not merely to discover a large number of valleys, but to adequately represent the statistics of those corresponding to the dominant low-energy (high-probability) regions of the landscape while also identifying spurious low-energy minima that influence the learning process. Conventional approaches to comparing a particular probability distribution to another (e.g., the one targeted during training) include the relative entropy \cite{goto_online_2023}, expectation values of selected observables under the given probability distribution \cite{gonzalez_testing_2021}, etc. The contribution of narrow LVs in those statistical comparisons is expected to be significantly smaller due to a smaller density of high-probability states compared to the contribution of broader LVs. It may not be critical to account for those states when comparing the probability distribution by classical means, and those LVs may not be present in an otherwise sufficiently representative sample. However, following the discussion above, most low-energy/high-probability LMs have a good chance of being near a TP. Therefore, the ability to sample those LMs is a critical part of the model’s learning of those TPs, even if the corresponding LV is narrow and does not contribute to a high density of highly probable states. 

It is our opinion that the discussion above justifies our focus on LMs and LVs when comparing the sampling performance of a QA and that of the classical Gibbs sampling, with a solid potential to reveal such performance differences as the ability to sample high-probability states from all or most of the low-energy LVs (including narrow ones), which at the later stages of the training means the ability to sample states corresponding to all or most of the TPs.
\subsection{Adiabatic Quantum Annealing---Ideal and Noisy}\label{sec2.2} 

The energy function of a general Ising spin glass model has the following form:

\begin{equation}
E(s) = \sum_{i=1}^{N-1} \sum_{j=i+1}^{N} J_{ij} s_i s_j - \sum_{j=1}^{N} h_j s_j,
\label{eq11}
\end{equation}
\noindent where $s_i$ is the spin of the qubits $i$, $J_{ij}$ are the strengths of the couplings between qubits $i$ and $j$, and $h_j$ is the qubit’s  $j$ local field. Since the RBM’s energy function $E(v, h)$ (Equation~(\ref{eq2})) is a particular case of the energy function of a general Ising spin glass, RBM embedding into the Chimera lattice would allow, generally speaking, using QA to find the lowest-energy state (or degenerate lowest-energy states) of $E(v,h)$. The reality of the QA hardware is such that it, instead, behaves like a sampler from the probability distribution describing the embedded model. 

In most RBM ML applications, the training and inference are performed at $T=1$ (e.g., see Equation~(\ref{eq1})). This is a shortcoming of using the D-Wave for RBM training, including sampling from the RBM probability distribution, because the D-Wave operates at a very low physical temperature. Since the primary purpose of a QA is the GS determination, ideally, only the GS would be returned by the D-Wave at the end of the adiabatic QA. The reality of the current QA hardware implementation is different. The following is responsible for QAs (and the D-Wave in particular) returning a multitude of excited states (ESs) in addition to (or sometimes instead of) the GS. 

The adiabatic approximation for a time-dependent Schr{\"o}dinger equation means that the system’s Hamiltonian is considered to change very slowly, ideally taking infinite time to evolve from its value at the initial time $t_0$ to its final value at $t_f$ (i.e., $H(t_0) \rightarrow H(t_f)$). The eigenvalues and associated eigenvectors of $H(t)$ also change slowly. If the system is in the state $\ket{E_j(t_0)}$ at the time $t = t_0$,  it will be in the state $\ket{E_j(t_f)}$ at $t = t_f$. Therefore, even though the $j${th}
 eigenstate changes, the system will always remain in the $j$th eigenstate $\ket{E_j(t)}$ of the changing Hamiltonian $H(t)$. This idea is behind adiabatic quantum computing. When the initial state of the system $\ket{E_0(t_0)}$ is the GS of its Hamiltonian $H(t_0)$, an adiabatic time development will lead to the GS of $H(t_f)$, even though the initial GS $\ket{E_0(t_0)}$ was very different from the searched final GS $\ket{E_0(t_f)}$. In the current implementation of the D-Wave, the qubits of the system are placed into an equal superposition of all $2^{n}$ possible states, which is the GS $\ket{E_0(t_0)}$ of a particular initial Hamiltonian (the so-called driver Hamiltonian) corresponding to an Ising spin glass with a transverse magnetic field. After that, the system is adiabatically changed from the driver to the problem Hamiltonian, in which case, it should remain at all times in the GS $\ket{E(t)}$ of the current Hamiltonian $H(t)$, and end up in the GS of the final targeted (problem) Hamiltonian $H(t_f)$ (e.g., in our case, the Hamiltonian (or the Energy Function) of the RBM under training or investigation).

Computations and visualizations of the time evolution of the GS and a few lowest eigenstates of an arbitrarily small problem with just a few qubits could be conducted even in a classroom setting to show how closely the energies of some of the eigenstates (and therefore the probabilities of observing those eigenstates) approach each other, even in an ideal case of infinitely slow evolution. This fact alone illustrates an easy deviation from the adiabatic theorem caused by the disappearance of a (required for adiabatic evolution) gap between the particular eigenvalue (the GS in the present context) and the rest of the Hamiltonian's spectrum, possibly resulting in this state escaping from the desirable pathway of remaining in the evolving GS of the evolving Hamiltonian. Anything less than infinitely slow evolution may result in a situation when the given $j$th state of the Hamiltonian $H(t^-)$ at an earlier time $t^-$ is not the $j$th state of the Hamiltonian $H(t^+)$ at a later time $t^+$, even when there is a gap between this eigenvalue and the rest of the Hamiltonian's spectrum. Other contributions to quantum fluctuations, as well as the finite temperatures of the QA hardware, additionally contribute to the probability of the final state being an eigenstate different from the GS of the final problem Hamiltonian. In practice, it could result in many (if not most) of the solution repetitions in a single D-Wave call (i.e., up to 10,000 solution repetitions allowed by the current version of the hardware) being different states from each other. Only small subsets of those states would be repetitions of the same states, which may or may not be the GS. And that multitude of states, instead of only one GS, is behind the idea of using a (often post-processed) set of D-Wave solution repetitions as a sample. 

Nothing in the quantum adiabatic theorem statement promises that deviations from the ideal adiabatic evolution should follow the Boltzmann probability distribution at any temperature. However, some of the previous studies \cite{benedetti_estimation_2016,marshall_power_2019} showed that the distribution of the D-Wave solutions follows the Boltzmann probability distribution at some low effective temperature, even though it is very different from the $T=1$ desirable for the RBM training. Regardless of that property, our interest in D-Wave sampling is to explore possibilities coming from quantum rather than thermal fluctuations to sufficiently frequently sample high-probability states in the RBM distribution that could, in practice, have very low probabilities of being sampled by classical methods, for example, because of the narrow basin of attraction \cite{koshka_comparison_2020}.

\section{Methodologies}\label{sec3}
\subsection{RBM Training by Classical CD-\textit{k}}\label{sec3.1}
In our previous work \cite{koshka_comparison_2020,koshka_toward_2020}, the OptDigits dataset was used for RBM training, with patterns additionally scaled down to $8\times7$ pixels. The original dataset was in greyscale [0--255], binarized, and converted to the Ising representation [$-$1, 1]. To achieve adequate embedding with the previous version of the D-Wave hardware (i.e., Chimera), which had lower connectivity, only 8 out of the 10 classes of the handwritten digits were used. In the present work, the same dataset was used but with a higher resolution; it was scaled down to either $12\times12$ or $8\times8$ pixels. The dataset was binarized similarly, but the Quadratic Unconstrained Binary Optimization (QUBO) representation was used instead of Ising, to make it closer to how RBMs are trained in classical ML. The RBM was trained using 1000 TPs. A contrastive divergence algorithm with $kG=5$ was used \cite{fischer_training_2014}. The previously developed modification of the weight-decay method \cite{koshka_toward_2020} was adopted to keep the RBM weights as small as possible during and after the training.
\subsection{RBM Embedding into the D-Wave Lattice}\label{sec3.2}
While the RBM was trained classically using the QUBO representation of Equation (2), the D-Wave hardware natively operates on the Ising representation of Equation (11). The trained QUBO-form model was submitted to the D-Wave directly via the Ocean SDK's DWaveSampler().sample\_qubo() function, which performs the QUBO-to-Ising conversion internally before execution on the QPU and converts the returned solutions back to QUBO form. In contrast to our previous work explicitly utilizing the Ising form of the RBM equations \cite{koshka_comparison_2020,koshka_toward_2020,koshka_comparison_2021}, no manual conversion between representations was performed by the authors. The D-Wave Advantage\_system4.1 \cite{dwave_architecture} uses an architecture called Pegasus, which is different from the previous version. As in the last version of the hardware, the qubits are connected to each other via couplers for which the user can specify the values of their strength $J_{ij}$. The Pegasus qubits are connected to up to 15 other qubits, compared to up to six connections in the Chimera topology used in our previous works \cite{koshka_comparison_2020,koshka_toward_2020,koshka_comparison_2021}. In addition to the couplers between qubits, each qubit has a bias ${h_j}$. As in the previous version, the D-Wave architecture with $N$ qubits represents a spin glass problem of $N$ spins given by the energy function shown in Equation~(\ref{eq11}). 

The actual graphs of interest to ML applications usually have much more than 15 couplers per logical unit (e.g., in the extreme case of the so-called fully connected BM, each logical unit must be connected to all the other units). Therefore, using each qubit as a logical unit and relying on D-Wave hardware couplers as RBM weights would not allow one to embed graphs with sufficient connectivity. Chaining multiple qubits to represent one logical unit of the graph is the primary approach to overcoming this problem. Once the graph with chained qubits is used to produce a D-Wave solution, the results must be post-processed to infer the value of each logical unit from the states of the qubits in the chain representing that logical unit. When not all the chained qubits in the D-Wave solution are aligned in the same direction (i.e., a so-called “chain break”), the D-Wave post-processing tools offer multiple approaches to determine the state of the logical unit, such as using a majority vote or discarding the sample with a chain break.  

This work used an RBM with 74 visible and 74 hidden units, as well as 154 visible and 72 hidden units. In contrast to our previous work, in which the qubits were manually selected for inclusion in each logical unit to achieve the best possible connectivity, the embedding into the Pegasus topology utilized the D-Wave’s minor-embedding method \cite{Dwave_minor-embedding_2024}. This method automatically handles the chaining of the qubits. Embedding the RBM manually is impractical when dealing with the more complex and denser layout of the Pegasus hardware.

The maximum values of $J_{ij}$ allowed by the D-Wave hardware range from $-1$ to $1$. Therefore, the abovementioned weight-decay method was used to keep the RBM weights in this range during the training described in Section~\ref{sec3.1}. Moreover, having some of the RBM weights equal or close to $1$ or $-1$ may interfere with the quality of the qubits chaining into a single unit. The use of ``ferromagnetic'' (i.e., ideally, having infinite strengths) couplings between the chained qubits ensures that the chained qubits all have the same value (i.e., aligned in the same direction). If the $J_{ij}$  representing the weights of the embedded RBM are comparable to the ``ferromagnetic'' bonds of the chained qubits, the ferromagnetic alignment may be violated. This problem is mitigated by scaling all the RBM weights after training, before embedding. However, this scaling could introduce its own problem for the quality of embedding the smallest weights since the D-Wave Quantum Processing Unit (QPU) has limited precision, and the realized small values may deviate from the intended values. Therefore, similar to our previous work, using qubit scaling after training before embedding was supplemented with the abovementioned effort to keep the weights of the trained RBM as small as possible.

\subsection{Use of the D-Wave for Classification, Pattern Reconstruction and Generation}\label{sec3.3}
After the RBM was trained classically using CD-\textit{k} \cite{fischer_training_2014} (Section~\ref{sec3.1}), the trained RBM was then embedded into the D-Wave QA (Section~\ref{sec3.2}), which was then used to classify and reconstruct missing regions of test images (hand-written digits). The test dataset comprised 300 patterns and completely differed from the training dataset.  The test patterns were selected randomly from the test dataset. In the reconstruction experiments, for each pattern, 54\% of the pixels were “clamped” in the D-Wave embedding, and the values of the other 46\%, including the labels, were sampled by the D-Wave. Clamping pixels into the D-Wave hardware was realized as follows. For the qubits corresponding to the pixels of the patterns that needed to be clamped, the local fields $h_j$ were set to the maximum possible value to ensure that this qubit value cannot be affected by other parameters of the Ising spin glass (i.e., by the values of the couplings). When the RBM unit (the pixel) needs to be clamped to $1$, the $h_j$ value of the corresponding qubit is set to $-4$. The $h_j$ value is set to 4 when the RBM unit needs to be clamped to 0. The rest of the local fields and couplers are set to values corresponding to the values of the embedding of the trained RBM. Following the procedure described in Section~\ref{sec3.2}, a scale factor was applied to the classical model before embedding the patterns. The D-Wave auto-scaling of couplers and biases was not used. A single D-Wave call can return 10,000 states (solution repetitions). The lowest-energy state is expected to provide the best reconstruction of the pixels and the class labels that were not clamped. However, that is not guaranteed to be the case since chain breaks (as described in Section~\ref{sec3.2}) may happen, resulting in the D-Wave solving an Ising spin glass model that is different from the intended one (i.e., somewhat different from the embedded RBM). As was mentioned in Section~\ref{sec3.2}, the D-Wave handles chain breaks automatically and offers users the choice to discard those samples or use majority vote to determine the value of the logical unit calculated from the values of qubits in the chain. However, this does not always give the correct result, especially when the weights and biases are not small enough compared to the ferromagnetic bonds in the chain of coupled qubits. Aggressive use of weight decay during the RBM training and scaling down of weights and biases before embedding, as described in Section~\ref{sec3.2}, significantly alleviated this problem in this work.

Finally, a similar procedure was applied when the goal was to generate an entire pattern instead of reconstructing a part of an image. However, following the abovementioned procedure, only those qubits representing the logical RBM units responsible for the class label were clamped. 
\subsection{Determination of the Local Valleys in the Energy Function}\label{sec3.4}
The classical search for the LVs (or, more precisely, the corresponding LMs) was conducted by starting from some initial states and performing a certain number of Gibbs sampling steps at temperature $T = 1$, followed by a relaxation to the bottom of the corresponding LV to find the LM. The relaxation was achieved by performing enough additional Gibbs sampling steps at temperature $T = 0$ (i.e., movement only downhill), which were terminated when the states of the RBM units stopped changing. When deciding which initial states to use, this work utilized an approach that is the most relevant for Gibbs sampling during CD-\textit{k} RBM training. In this approach, each of the 1000 TPs was
 used as an initial state for a Gibbs sampling chain. Essentially, this procedure finds LVs to which the states sampled during a CD-\textit{k}-based training belong. A bitwise comparison was used to keep only distinct LVs on the list of found LVs. 

For the LV search using the D-Wave, the goal was similar---find LVs to which the states returned by all 10,000 (or, in some cases, a smaller number) of solution repetitions in the given D-Wave job belong. Prior to this relaxation step, any solution repetitions exhibiting chain breaks (Section \ref{sec3.2}) were repaired using majority vote; no samples were discarded on account of chain breaks. Consequently, the $N_{\text{LV}}$ values and RBM energy histograms reported in this work
 reflect the complete set of D-Wave solution repetitions. The classically trained RBM model was embedded, as described in Section~\ref{sec3.2}. Next, we started from each of the distinct states returned by the D-Wave after the specified number of solution repetitions. A classical Gibbs sampling chain was conducted at $T = 0$ to achieve relaxation to the bottom of the LV in which the given D-Wave solution state was residing. A bitwise comparison was used to keep only distinct LVs on the list of found LVs, as was done in the classical LV search.

In Section~\ref{sec4.4}, the LVs found by the classical Gibbs sampling and by the D-Wave are analyzed with respect to the RBM energy. The energy of the states of the found LMs was calculated according to Equation~(\ref{eq2}).

The D-Wave supports two annealing schedules: forward (standard) and reverse annealing. Custom annealing schedules can be used, which may include a pause or quench \cite{dwave_annealing}. However, only forward annealing was used in all the experiments conducted in this work. In some experiments (see Section~\ref{sec4.2}), the annealing time was varied from 1 {$\upmu$}s 
 to its maximum value (2000 {$\upmu$}s). Based on the results of Section~\ref{sec4.2}, the default annealing time of 20 {$\upmu$}s was optimal for this work and was used in most of the reported experiments unless specified otherwise.
\section{Results and Discussion}\label{sec4}
\subsection{The Quality of Embedding the Trained RBM into the D-Wave Lattice}\label{sec4.1}
\begin{figure}[H]
\includegraphics[]{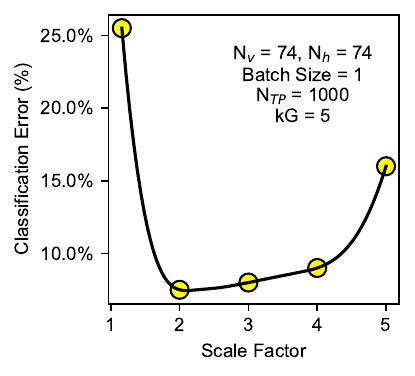}%width=4.0 cm
\caption{The RBM classification error as a function of the scale factor (SF). The RBM had $N_v = 74$ and $N_h = 74$; it was trained with 1000 TPs from the OptDigits dataset. The D-Wave annealing time of 20 {$\upmu$}s was used. As the figure shows, the optimal value of the SF that yields the best classification error is close to 2. This value of the SF was used to embed all the RBM models throughout this paper in D-Wave.\label{fig1}}
\end{figure}

The classically trained RBM had 74 visible and 74 hidden units; it was trained with 1000 patterns from the OptDigits dataset. The D-Wave solver (Advantage\_system4.1) used in this work allowed using ferromagnetic couplers to chain qubits into a given logical unit with the maximum value of $-1$. As described in Section~\ref{sec3.2}, a significant effort was made to keep the RBM weights $\omega_{ij}$ sufficiently smaller than the maximum (i.e., $-1$) strengths $J_{ij}$ of the ferromagnetic couplers, which was accomplished by employing the modified weight-decay procedure and using an optimal scale factor (SF). The optimal value of the SF was determined by examining the dependence of the classification error of a fully trained RBM on SF (Figure~\ref{fig1}). 

\vspace{-6pt}

The value of the SF that yields the best classification error was found to be close to 2. This value of the SF was used when embedding all the instances of the trained RBMs in this work in the D-Wave hardware. When using the D-Wave for classification, the test patterns were clamped by assigning the maximum negative/positive allowed values of the local fields ($h_j= \pm 4$) to the qubits corresponding to the RBM logical units representing the pixels of the image. From 10,000 D-Wave solution repetitions in a single D-Wave call, the values of the labels were obtained from the corresponding qubits of the lowest-energy state (i.e., the GS).

It should be noted that investigating the SF dependence of the classification error in this work was purely practical---to improve the embedding quality. Investigation of the mechanisms of the observed dependence was beyond the scope of this work.

\begin{figure}[H]
\hspace{-6pt} \includegraphics[]{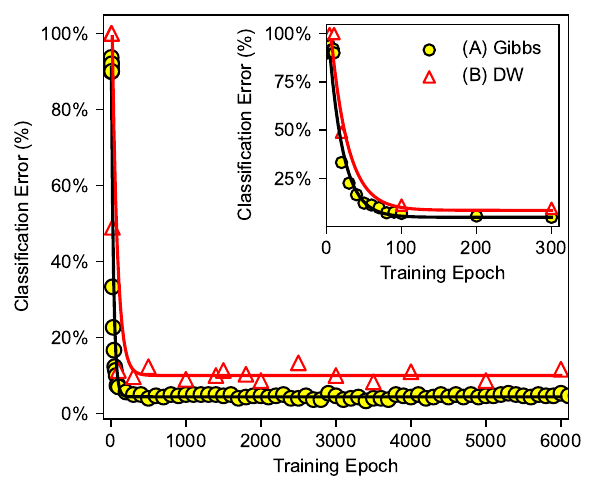}
  \caption{The classification error of the classically trained RBM as a function of the training epoch: (A) using the Gibbs sampling to reconstruct the labels, and (B) using the GS of the qubits that represent the classification labels. The inset in Figure~\ref{fig2} shows a magnified part of the same dependence at the first stages of the training. These results showed that the RBM energy function of the D-Wave embedding is sufficiently close to that of the original RBM used for the embedding.\label{fig2}}
\end{figure}

Next, the RBM classification error for the D-Wave-based reconstruction of the labels was compared to the classification error obtained with the classical Gibbs sampling.  The RBM classification error as a function of the training epoch is shown in Figure~\ref{fig2}, with Curve (A) and Curve (B) corresponding to the classical and the D-Wave-based reconstructions, respectively.  
In both cases, the RBM had been trained classically. The RBM model in Figure~\ref{fig2} was the same as in Figure~\ref{fig1}. The classical reconstruction in Curve (A) was conducted conventionally. A testing pattern with randomized labels was used to initiate the MC chain. The RBM units corresponding to the pixels of the image were clamped, while the state of the label-units was updated after each Gibbs sampling step. After 400 MC steps were completed, the states of the label-units after each of the 100 additional MC steps were used to obtain the majority vote for the value of the label. For the D-Wave-based classification in Curve (B), the same RBM embedding into the D-Wave lattice and the same label-reconstruction procedure as in Figure~\ref{fig1} were used. 

\vspace{-6pt}
%Fig 2

It follows from the figure that the quality of the embedding was sufficient to achieve the classification error with the D-Wave that is sufficiently close (for our purposes) to that obtained by the classical Gibbs samplingreconstruction, which serves as evidence that the RBM energy function of the D-Wave embedding is close to that of the original RBM used for the embedding.

\begin{figure}[H]

\subfloat[\centering]{\includegraphics[]{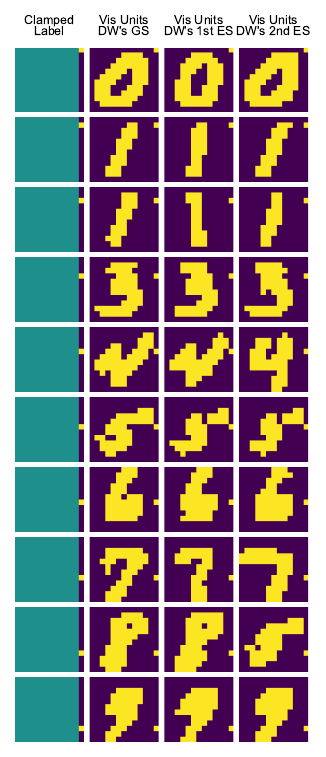}}
%\hfill
\subfloat[\centering]{\includegraphics[]{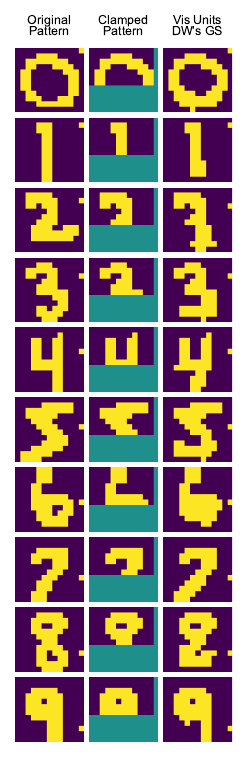}}

\caption{(\textbf{a}) Examples of using the D-Wave for image generation; (\textbf{b}) successful examples of partial image and label reconstruction (classification). The first column shows the values of the clamped label. The rest of the columns show the three lowest-energy states of the visible units found by the  D-Wave in ascending order of the RBM energy from left to right. In (\textbf{b}), qubits corresponding to approximately 54\% of the image pixels were clamped, and the D-Wave found the rest of the qubits to reconstruct the missing part of the image. The left column in (\textbf{b}) shows the original test patterns, the middle is the corresponding patterns of D-Wave embedding showing which pixels are clamped, and the right is the result of the reconstruction from the D-Wave. Pixels are shown using a three-color scheme: yellow indicates a unit clamped to 1, purple/dark blue indicates a unit clamped to 0, and teal indicates a free (unclamped) unit whose value is sampled by the D-Wave.\label{fig3}}
\end{figure}

As a qualitative illustration of the adequacy of the embedding, the D-Wave was applied to the classically fully trained RBM to perform image generation and reconstruction tasks. A bigger RBM, with $N_v =154$ and $N_h = 72$, was used in this work to demonstrate the capability of the Pegasus D-Wave hardware to embed big graphs. The D-Wave annealing time used was 20 {$\upmu$}s. Figure~\ref{fig3}a shows examples of using the D-Wave for image generation. Figure~\ref{fig3}b shows successful examples of using the D-Wave to reconstruct a partial image and the corresponding label (i.e., classification).  In (a), the qubits representing the logical one-hot labels were assigned the highest ($h_j= \pm4$) value of the local field to clamp the label to the desired value. The rest of the qubits had the values of the local fields determined by the RBM embedding after applying the optimal value of the SF ($SF = 2$) from Figure~\ref{fig1}. The first column in Figure~\ref{fig3}a shows the clamped label. The rest of the columns show the three lowest-energy states of the visible units found by the D-Wave, arranged from left to right in ascending order of the RBM energy. A reasonably good performance of the RBM embedding as a generative model has been achieved.  The other, smaller graph size used in this work ($N_v =74$ and $N_h = 74$) has shown comparable generation and reconstruction capabilities. 
\vspace{-16pt}
%Fig 3

%\newpage

In Figure~\ref{fig3}b, qubits corresponding to approximately 54\% of the image pixels were clamped, and the rest of the qubits took values from the lowest energy state out of 10,000 solution repetitions, to reconstruct the missing part of the image. The clamping and reconstruction were done as in Figure~\ref{fig3}a. The left column in (b) shows the original test pattern, the middle column is the pattern of the D-Wave embedding showing which pixels are clamped, and the right column shows the result of the reconstruction from the D-Wave solution. The label value determined this way was used to calculate the classification errors in Figure~\ref{fig2}. However, in Figure~\ref{fig2}, the classification labels and the error were determined while clamping the entire image rather than a portion of the image.
\subsection{Finding the GS Versus Other Local Valleys}\label{sec4.2}
One of the intuitive expectations about a desirable behavior of the D-Wave as a sampling engine is not only, or not even necessarily, the requirement that the sampled states follow a Boltzmann distribution (especially if those states are to be used as seeds for subsequent MC). More important is its ability to return a set of states that belong to a wide range of different LVs, adequately representing the complexity of the configuration space. In this regard, as has been mentioned in the Introduction, there is a logical concern about the previously reported trend of the newer generations of the D-Wave hardware to improve the so-called time to solution (TTS) for finding the GS (which is the primary job of a QA) \cite{pelofske_comparing_2023}. A concern is that the TTS improvement may also cause the reduced ability to obtain sufficiently large samples that have high variance. While investigation of the tradeoff between TTS and the sample size is beyond the scope of this work, we paid some attention to this tradeoff, specifically with respect to the objective of establishing if any adjustments in the QA conditions could benefit the sampling objectives.

\begin{figure}[H]
\begin{adjustwidth}{-\extralength}{0cm}
\centering
\subfloat[\centering]{\includegraphics[]{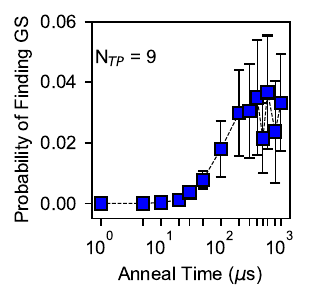}}
%\hfill
\subfloat[\centering]{\includegraphics[]{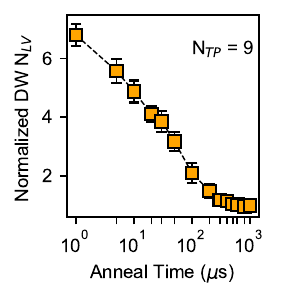}}
\subfloat[\centering]{\includegraphics[]{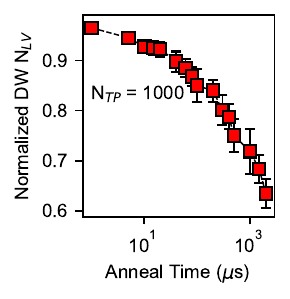}}
%\hfill
\end{adjustwidth}
\caption{The effect of D-Wave annealing time on two expectedly conflicted outcomes: (\textbf{a}) the probability of the D-Wave finding the ground state (GS) and (\textbf{b}) the average per D-Wave run number of distinct LVs found by the D-Wave, normalized to the number of training patterns (nine in this case). Both (\textbf{a},\textbf{b}) correspond to the embedding of a “simple” RBM-training case (different from Figures~\ref{fig1}--\ref{fig3})---the RBM trained with only nine $8\times8$ patterns from the OptDigits dataset. The improved ability of the D-Wave to find the GS correlates with the diminished ability to find distinct LVs (i.e., the reduced variety of the sample).  (\textbf{c}) Similar to (\textbf{b}) but for the main RBM training case investigated in this work, RBM trained with 1000 training patterns (as in Figures~\ref{fig1}--\ref{fig3}).\label{fig4}}
\end{figure}

The D-Wave annealing time is the main parameter influencing the probability of finding the GS ($P_{\text{GS}}$). Unfortunately, the D-Wave provided a very low frequency of finding the GS from 1000 (or even 10,000) solution repetitions, for such a complex energy landscape as the RBM energy function (for the fully trained RBM as well as at earlier training epochs). In some cases, only one of 10,000 solutions, or none at all, is the GS. An impractically greater number of expensive D-Wave calls would be required to obtain sufficient statistics for such an energy landscape to calculate $P_{\text{GS}}$. Therefore, in the following experiment, an RBM trained with only nine $8\times8$ patterns from the OptDigits dataset was investigated to determine $P_{\text{GS}}$ and compare it to some measure of sample variance. The two expectedly conflicting characteristics were investigated---$P_{\text{GS}}$ (Figure~\ref{fig4}a) and the average number of distinct LVs $N_{\text{LV}}$ found by the D-Wave per run (Figure~\ref{fig4}b).
 
\begin{equation}
    P_{GS} = \frac{\text{Number of times GS appeared in } N_{\text{smp}}}{N_{\text{smp}}} ,
\label{eq12} 
\end{equation}

$N_{\text{LV}}$ is shown normalized to the number of TPs. In Figure~\ref{fig4}a, at each anneal time, one D-Wave call ($N_{\text{DW}} = 1$) with 1000 D-Wave solution repetitions ($N_{\text{sol}} =$ 1000) was performed to obtain 1000 samples ($N_{\text{smp}} = N_{\text{DW}}  \times N_{\text{sol}}$) and calculate the $P_{\text{GS}}$ according to Equation~(\ref{eq12}) \cite{stollenwerk_quantum_2020}. This process was repeated ten times, and the average value of $P_{\text{GS}}$ was calculated. As expected, the probability of finding the GS initially improved with the annealing time up to approximately 200 {$\upmu$}s. Further increases in the annealing time did not cause any improvement for this simple RBM case. 

In Figure~\ref{fig4}b, $N_{\text{LV}}$ for LVs found by the D-Wave is normalized to the number of TPs ($N_{\text{TP}}$), which was nine in this simple example. As follows from the figure, at least for this case of a simple energy landscape (i.e., expectedly small $N_{\text{LV}}$ due to the small $N_{\text{TP}}$), the improved ability of the D-Wave to do its primary job, find the GS, correlates with a diminished ability to find distinct LVs. Therefore, in general, shorter annealing times could be helpful when obtaining a higher number of LVs from a D-Wave call is desirable. It is important to note that the nine-pattern RBM was used in this part of our work specifically because of its small dimensionality, with a tractable energy landscape making $P_{GS}$ statistically measurable within a practical number of D-Wave calls. Such measurement is not feasible for the more complex, 1000-pattern RBM used throughout the rest of this work. The tradeoff involved with this choice should therefore be interpreted as a qualitative validation of the expected trend under simplified, tractable conditions, rather than as a quantitative prediction. As shown next, this promising trend did not translate into a practical advantage in $N_{\text{LV}}$ for the more complex RBM.

\vspace{-8pt}
%Fig 4

Figure~\ref{fig4}c shows the annealing time dependence for the D-Wave-found $N_{\text{LV}}$ for the RBM trained with $N_{\text{TP}} =$ 1000, the training case of Figures~\ref{fig1}--\ref{fig3}. For this main RBM case investigated in this work, an attempt to vary the annealing time from the default value (20 {$\upmu$}s) has not produced a desirable substantial increase in the $N_{\text{LV}}$ found by the D-Wave, which means no significant increase in the variance of the “sample” from the D-Wave was achieved.  

As mentioned earlier, $P_{\text{GS}}$ (which was determined in Figure~\ref{fig4}a) could not be calculated for this case to compare to $N_{\text{LV}}$, since an impractically greater number of expensive D-Wave calls would be required to obtain sufficient statistics for such an energy landscape.

\subsection{Comparison of $N_{\text{LV}}$ Found by the D-Wave and by the Classical Gibbs Sampling}\label{sec4.3}
As mentioned in the introduction, the focus of this work was to establish how many, and what kind of, LVs a sample from an RBM probability distribution belongs to when sampled by the D-Wave, in comparison to those obtained by classical Gibbs sampling, specifically at conditions relevant for RBM training by CD-\textit{k}.

First, it was investigated how $N_{\text{LV}}$ during the classical sampling depends on \textit{k}G within the range of typical \textit{k}G values. Therefore, an LV search with three different Gibbs schedules was conducted: $kG=1$, $kG=10$, and $kG=100$. In each \textit{k}G schedule, repeated Gibbs sampling chains were used to produce 10,000 samples, and then it was determined how many LVs all those samples belonged to, following the procedure described in Section~\ref{sec3.4}. The Gibbs chains were initiated from each of the 1000 TPs, followed by the \textit{k}G number of Gibbs steps at temperature $T = 1$ and the subsequent relaxation to the bottom of the LV at $T=0$. The desired number of samples $N_{\text{smp}} =$ 10,000 was obtained using the formula $N_{\text{smp}} = N_{\text{TP}}\times N_{\text{rpt}}$, where $N_{\text{TP}} =$ 1000 is the number of TPs and $N_{\text{rpt}} = 10$ is the number of repetitions of the Gibbs chains described above. 

\begin{figure}[H]
\begin{adjustwidth}{-\extralength}{0cm}
\centering
\subfloat[\centering]{\includegraphics[]{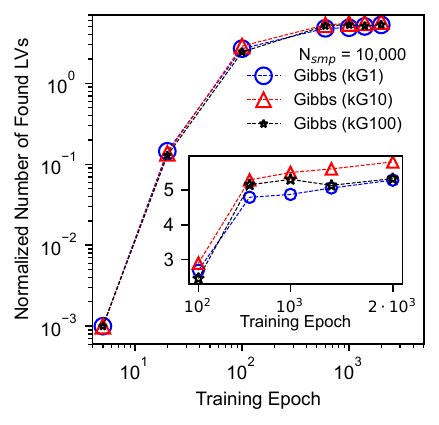}}
%\hfill
\subfloat[\centering]{\includegraphics[]{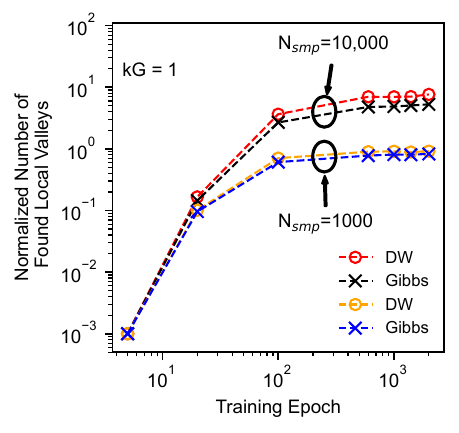}}
\end{adjustwidth}

  \caption{The number of the LVs $N_{\text{LV}}$ normalized to the number of training patterns (1000), shown as a function of the training epoch. (\textbf{a}) $N_{\text{LV}}$ of LVs found by the classical Gibbs sampling for three different values of \textit{k}G (1, 10, and 100). $N_LV$ only slightly increases with the number of Gibbs jumps. The inset in Figure (\textbf{a}) shows a magnified view of the same dependence starting from 90 epochs, when the classification error drops below 8\%. (\textbf{b}) $N_{\text{LV}}$ of LVs found by the D-Wave and by the classical Gibbs sampling with $kG= 1$, when both were used to generate the same number of (not necessarily distinct) samples $N_{\text{smp}}$. Two cases are shown: $N_{\text{smp}} =$ 1000 and $N_{\text{smp}}=$ 10,000 for both the D-Wave and the classical sampling.\label{fig5}}
\end{figure}

In Figure~\ref{fig5}a, $N_{\text{LV}}$ found by the three different schedules ($kG = 1$, 10, and 100) of Gibbs sampling is shown as a function of the training epoch. Of course, only distinct LVs were counted. $N_{\text{LV}}$ is plotted normalized to $N_{\text{TP}}$ (1000). It can be observed that the number of classically found LVs may slightly increase with \textit{k}G in some range of the \textit{k}G values, which is tentatively explained by the expectation that a higher number of Gibbs steps makes it more likely to reach equilibrium. However, the difference is relatively slight (7\% between the two extremes corresponding to $kG=1$ versus $kG=100$). From Figure~\ref{fig5}a, $N_{\text{LV}}$ for $kG=10$ is no more than 10\% higher than that for $kG=1$ and for $kG=100$, the difference is even more negligible. For that reason, and since $kG =1$ is frequently used in classical RBM training, unless specified otherwise, some of the subsequent comparisons of the D-Wave and the Gibbs sampling of this work, especially when the Gibbs sampling LV search was costly, were conducted for $kG=1$ only.

\vspace{-16pt}

%Fig 5

Next, $N_{\text{LV}}$ found by Gibbs sampling with $kG=1$ was compared to that found by the D-Wave (Figure~\ref{fig5}b). Both Gibbs sampling and the D-Wave were used to generate the same number of (not necessarily distinct) samples $N_{\text{smp}}$. Two cases are shown in Figure~\ref{fig5}b: $N_{\text{smp}}=$ 1000 and $N_{\text{smp}}=$ 10,000 for both the D-Wave and the classical sampling. For the D-Wave, the desired $N_{\text{smp}}$ was generated by obtaining $N_{\text{sol}}$ solution repetitions from a single D-Wave call, with two values of $N_{\text{sol}}$: 1000 and 10,000. For the classical Gibbs sampling, the desired $N_{\text{smp}}$ (1000 and 10,000 in this experiment) was obtained with $N_{\text{smp}} = N_{\text{TP}} \times N_{\text{rpt}}$, where $N_{\text{TP}}=$ 1000 and $N_{\text{rpt}}$ was 1 and 10.

While this may not always be the case for other RBM training schedules, the particular RBM training conditions used in this work (constrained by the need to make the RBM suitable for D-Wave embedding) produced the following trend. Figure~\ref{fig5}b shows that $N_{\text{LV}}$ steadily increases with the training epoch. When both techniques acquire the same $N_{\text{smp}}$, the samples produced by the D-Wave were found to belong to a somewhat greater number of LVs than the classical samples. For example, when $N_{\text{smp}}=$ 10,000, after 1000 RBM training epochs, $N_{\text{LV}}$ from the D-Wave was 43.5\% higher than $N_{\text{LV}}$ from the Gibbs sampling. At earlier epochs, the difference was less significant (e.g., 14.5\% difference after 20 epochs). For $N_{\text{smp}} =$ 1000, the difference was more negligible, 12.9\% and 5.2\% at 1000 and 20 epochs, respectively.

\begin{figure}[H]

\subfloat[\centering]{\includegraphics[]{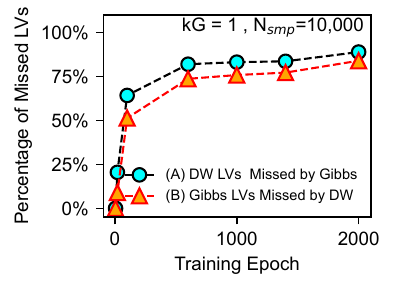}}
\subfloat[\centering]{\includegraphics[]{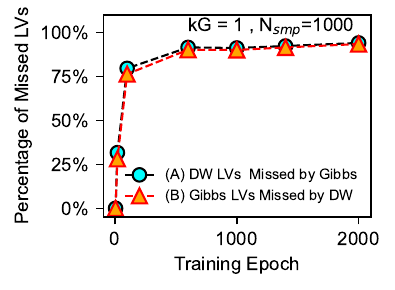}}\\
\subfloat[\centering]{\includegraphics[]{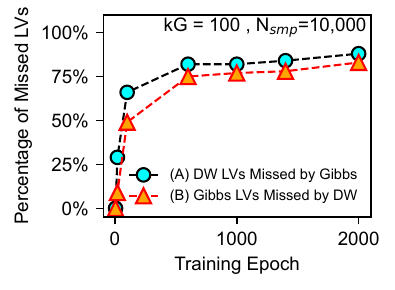}}
\subfloat[\centering]{\includegraphics[]{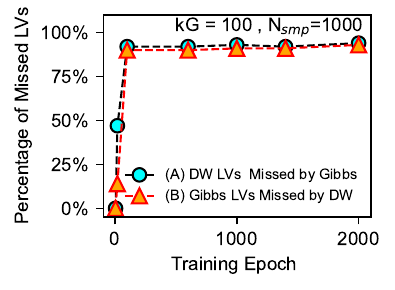}}
  \caption{Percentage of LVs found by one technique but missed by the other.  Curve (A): the percentage of D-Wave-found LVs missed by the classical Gibbs sampling. Curve (B): the percentage of Gibbs sampling-found LVs missed by the D-Wave. (\textbf{a}) $kG=1$, and the number of samples obtained from both the D-Wave and Gibbs sampling was $N_{\text{smp}} =$ 10,000. (\textbf{b}) The same as (\textbf{a}) but with $N_{\text{smp}} =$ 1000 for both the D-Wave and classical Gibbs sampling.  For the smaller $N_{\text{smp}}$, the overlap of LVs found by the two techniques was even smaller than for the larger $N_{\text{smp}}$. {(\textbf{c},\textbf{d}): the same as (\textbf{a},\textbf{b})}, but $kG=100$. The percentage of misses by either technique of the LVs found by the other technique is not much different between $kG=100$ and $kG=1$. For all the cases, the number of LVs missed by the other technique was smaller at early stages of training. The misses by the Gibbs sampling were somewhat higher than those by the D-Wave.\label{fig6}}
\end{figure}

As was reported in References~\cite{koshka_comparison_2020,koshka_toward_2020} for the Chimera graph of the D-Wave hardware, in addition to the different numbers of LVs to which the D-Wave samples and the Gibbs samples belong, many of those LVs revealed by the two techniques are different. Therefore, the next question was to establish how many of the LVs found by one technique are also seen by the other technique and how many are missed by the other technique. For example, for $N_{\text{smp}} =$ 1000 (which is the number of samples acquired during a single epoch of a CD-\textit{k} training with 1000 TPs), it would not be surprising to see the two sets of LVs not overlap much, even if produced by two equally capable sampling techniques. On the other hand, when more samples are acquired (e.g., $N_{\text{smp}}=$ 10,000), the expectation is that sufficient statistics should be acquired, making most of the “important” LVs more likely to overlap for any two suitable sampling methods that do similarly good jobs. This speculation is verified (and is found mostly incorrect) in Figure~\ref{fig6}. 

\vspace{-12pt}
%Fig 6

In Figure~\ref{fig6}, the percentage of the LVs found by one technique that are missed by the other technique is shown as a function of training epochs. In (a--d)
, curve (A) is for the percentage of the D-Wave-found LVs missed by the Gibbs sampling, and curve (B) is for the percentage of the Gibbs-found LVs missed by the D-Wave. $kG =1$ in the top two figures. \textit{k}G is increased to 100 in the bottom two figures. $N_{\text{smp}}=$ 10,000 in the two figures on the left. $N_\text{{smp}} $ is reduced to 1000 in the two figures on the right. From Figure~\ref{fig6}a, after 600 training epochs, more than 80\% (90\% at 2000 epochs) of the D-Wave LVs were missed by the Gibbs sampling. In turn, more than 70\% of Gibbs-found LVs were missed by the D-Wave. For both curves, the number of LVs missed by the other technique was smaller at the early stages of training (e.g., after 20 epochs, the Gibbs sampling missed 20\% of the D-Wave LVs and 9\% of Gibbs LVs were missed by the D-Wave).

When using a smaller $N_\text{{smp}} $ of 1000 (Figure~\ref{fig6}b), while $N_{\text{LV}}$ found by both techniques (Figure~\ref{fig5}) was smaller, the overlap of LVs found by the two methods was even smaller than for the larger $N_{\text{smp}} $. From Figure~\ref{fig6}b, after 600 training epochs, more than 91.5\% (94\% at 2000 epochs) of the D-Wave LVs were missed by the Gibbs sampling. The percentage of classically found LVs missed by the D-Wave was similar (90\% at 600 epochs and 93\% at 2000 epochs).
 
As was observed in Figure~\ref{fig5}, for the higher $kG=100$, the difference between $N_{\text{LV}}$ found by the two techniques is close to that for $kG=1$. However, the overlap (or the difference) between the two sets of LVs (the D-Wave-found and the Gibbs-found) may more dramatically depend on \textit{k}G of the classical search, even if $N_{\text{LV}}$ does not. Therefore, the experiments of Figure~\ref{fig6}a,b were repeated at $kG=100$. From Figure~\ref{fig6}c,d, it can be concluded that the percentage of misses by either technique of the LVs found by the other method is close to that at $kG=1$.

\subsection{$N_{\text{LV}}$ Distribution by RBM Energies}\label{sec4.4}
The final important question of this work was about what kind of LVs found by one of the two sampling techniques are missed (or not missed) by the other technique. More specifically, we wanted to analyze missed and not-missed LVs with respect to the energy (and therefore the probability) of the states of the corresponding LMs.

\begin{figure}[H]

\subfloat[\centering]{\includegraphics[]{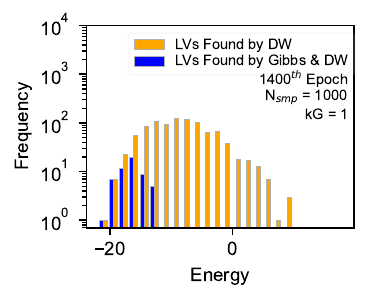}}
\subfloat[\centering]{\includegraphics[]{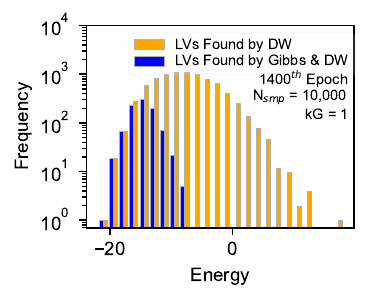}}\\
\subfloat[\centering]{\includegraphics[]{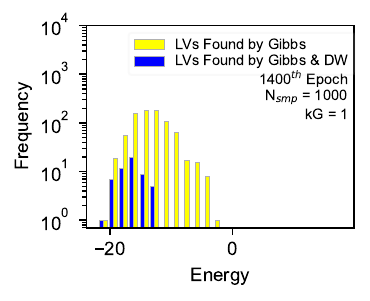}}
\subfloat[\centering]{\includegraphics[]{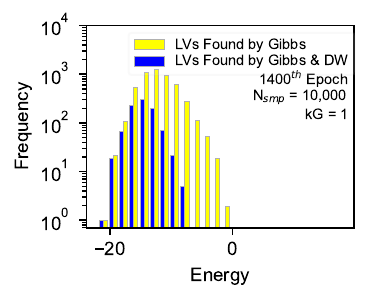}}

  \caption{(\textbf{a},\textbf{b}): Histograms of the RBM energies of all the LMs found by the D-Wave (orange) and similar histograms for only the portion of those LMs that coincide with LMs found by the classical Gibbs sampling~(blue). (\textbf{c},\textbf{d}): Histograms of the RBM energies of all the LMs found by classical Gibbs sampling (yellow) and similar histograms for only the portion of those LMs that coincide with LMs found by the D-Wave (blue). All the histograms are for the 1400th epoch and $kG=1$. For (\textbf{a},\textbf{c}), the number of samples $N_{\text{smp}}=$ 1000 for both Gibbs sampling and the D-Wave. For (\textbf{b},\textbf{d}),  $N_{\text{smp}}=$ 10,000 for both Gibbs sampling and the D-Wave.  In all the cases, each technique misses a substantial number of LVs found by the other. While the misses are the most pronounced for the high-energy (low-probability) states, a considerable number of misses can also be observed for intermediate energies (probabilities) and even for relatively low energies (high probabilities).\label{fig7}}
\end{figure}

\begin{figure}[H] 

\subfloat[\centering]{\includegraphics[]{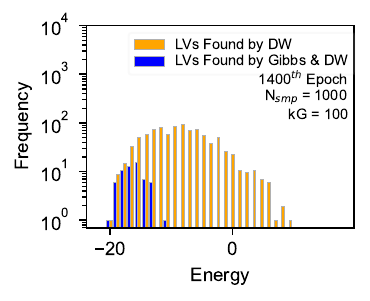}}
\subfloat[\centering]{\includegraphics[]{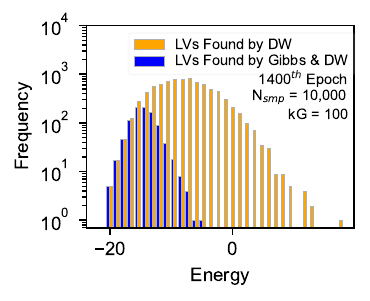}}\\
\subfloat[\centering]{\includegraphics[]{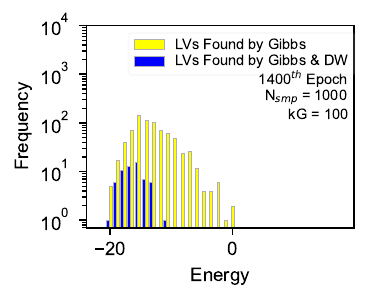}}
\subfloat[\centering]{\includegraphics[]{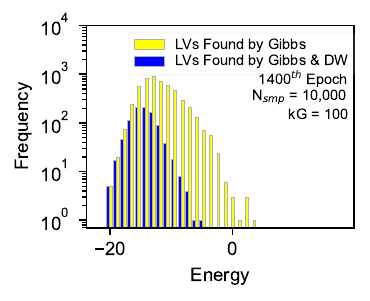}}

  \caption{As
 in Figure~\ref{fig7}, but $kG =100$ instead of $kG = 1$ during the Gibbs sampling search. All histograms are still for the 1400th epoch. Similar to Figure~\ref{fig7}, both Gibbs sampling and the D-Wave find most of the same low-energy (high-probability) LVs (a significant overlap of the LVs found by Gibbs sampling and D-Wave for the low-energy (high-probability) LMs). Figure~\ref{fig5}b shows that the D-Wave finds somewhat more LVs than Gibbs sampling. However, many of those LVs found by the D-Wave but missed by the Gibbs sampling are low-probability (high-energy) states.\label{fig8}}
\end{figure}

\begin{figure}[H]

\subfloat[\centering]{\includegraphics[]{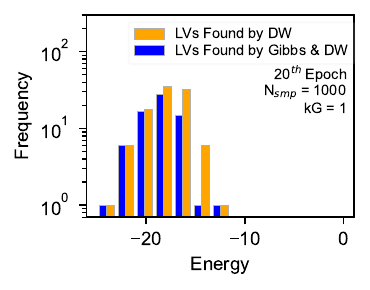}}
\subfloat[\centering]{\includegraphics[]{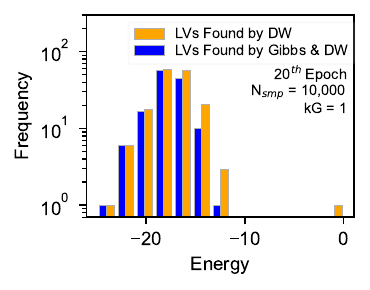}}\\
\subfloat[\centering]{\includegraphics[]{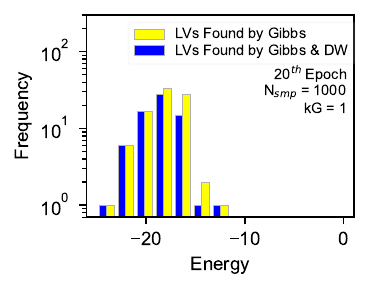}}
\subfloat[\centering]{\includegraphics[]{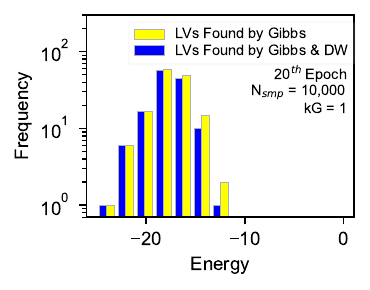}}

  \caption{As
 in Figure~\ref{fig7}, but after 20 training epochs instead of 1400 training epochs. At this early stage of training, the number of found LVs is much smaller than that at later stages of training (e.g., for $N_{\text{smp}}=$ 10,000, 145 LVs were found by Gibbs sampling after 20 epochs compared to 5059 LVs after 1400 epochs (see Figure~\ref{fig5}b)). As was observed in Figure~\ref{fig6}a, after 20 epochs, a large percentage of LVs found by either technique are also found by the other. This is confirmed by slight differences in the dark and light histograms in this figure. However, even though the histograms overlap entirely for the most low-energy (high-probability) LMs, in the intermediate- and high-energy parts of the spectrum, both the D-Wave and Gibbs sampling miss a non-negligible percentage of LVs found by the other technique.\label{fig9}}
\end{figure}

\begin{figure}[H]

\subfloat[\centering]{\includegraphics[]{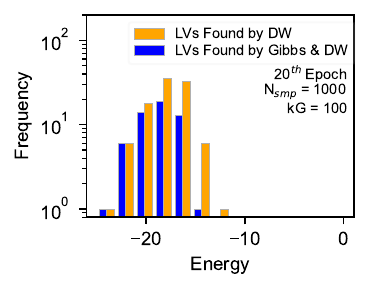}}
\subfloat[\centering]{\includegraphics[]{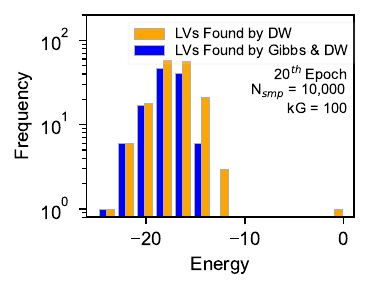}}\\
\subfloat[\centering]{\includegraphics[]{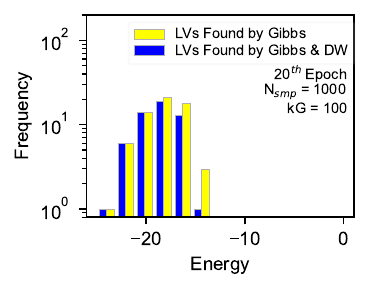}}
\subfloat[\centering]{\includegraphics[]{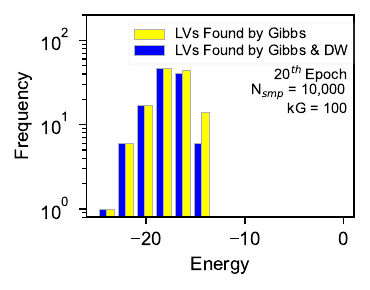}}

   \caption{As in Figure~\ref{fig9}, but $kG =100$ instead of $kG = 1$ during the Gibbs sampling search. At this early stage of training, the results for $kG=100$ (this figure) are like those for $kG=1$ (Figure~\ref{fig9}).\label{fig10}}
\end{figure}

The remaining comparisons of the D-Wave and the Gibbs sampling in this work (Figures~\ref{fig7}--\ref{fig10}) are presented in the form of histograms of the RBM energies of all the LMs found by one of the two

%Fig 7 
\vspace{-12pt}

\noindent techniques (orange or yellow) and similar histograms for only the portion of those LMs that coincide with LMs found by the other technique (blue). In all four figures (Figures~\ref{fig7}--\ref{fig10}), (a,b)
 show histograms for the LMs found by the D-Wave (orange), along with histograms for only the portion of those LMs that coincide with LMs found by Gibbs sampling (blue). Similar but opposite, in all four figures, (c,d) are histograms for the LMs found by Gibbs sampling (yellow) and similar histograms for only the portion of those LMs that coincide with LMs found by the D-Wave (blue). 

Figures~\ref{fig7} and~\ref{fig8} show the RBM trained with 1400 epochs. The difference between the two figures is that the classical LV search utilized $kG=1$ in Figure~\ref{fig7} and $kG=100$ in Figure~\ref{fig8}. Figures~\ref{fig9} and \ref{fig10} show the 20th epoch, using $kG=1$ and $kG=100$ for Figure~\ref{fig9} and Figure~\ref{fig10}, respectively.

\vspace{-12pt}
%Fig 8 

An additional difference explored in Figures~\ref{fig7}--\ref{fig10} is in $N_{\text{smp}} $, which was always the same for the D-Wave and Gibbs sampling but changed between some figures. Specifically, for (a,c) (i.e., left), the number of samples $N_{\text{smp}}=$ 1000 for both Gibbs sampling and the D-Wave. For (b,d) (i.e., right), the number of samples $N_{\text{smp}}=$ 10,000, also for both Gibbs sampling and the D-Wave.

For the case of 1400 training epochs (Figures~\ref{fig7} and \ref{fig8}), both Gibbs sampling and the D-Wave find most of the same low-energy/high-probability LVs, which is evidenced by the high overlap of the left portion of the histograms found by Gibbs sampling and the D-Wave, for both $N_{\text{smp}}=$ 10,000 and $N_{smp}=$ 1000 cases. We had already established that the D-Wave finds a higher number of LVs than Gibbs sampling~(recall Figure~\ref{fig5}b). Now, we learn from Figures~\ref{fig7} and \ref{fig8} that many of the LVs found by the D-Wave but missed by Gibbs sampling are high-energy/low-probability states.

The greater number of samples (Figures (b,d)
) caused a more substantial overlap between the classical and the D-Wave samples compared to Figures (a,c) for all four cases (Figures~\ref{fig7}--\ref{fig10}). However, even for $N_{\text{smp}}=$ 10,000 ((b,d)), which is $10\times$ more than the number of samples acquired for a single RBM training epoch, each technique misses a substantial number of LVs found by the other (orange histograms not overlapping with blue), when the RBM energy function is analyzed after 1400 epochs (Figures~\ref{fig7} and \ref{fig8}).

\noindent While
 the misses in Figure~\ref{fig7} are the most pronounced for the high-energy/low-probability states, many misses can also be observed for intermediate energies/probabilities and even for relatively low energies/high probabilities. In Figure~\ref{fig8}, which used $kG=100$ (also for the RBM trained with 1400 epochs), a qualitatively very similar behavior is produced, with minor quantitative differences. 

It should be noted that while this kind of difference between any (even considerably similar) sampling techniques would not be surprising when samples from a probability distribution are considered, our comparison is for LVs. A modest number (when normalized to the number of TPs) of high-probability LVs is expected from an RBM, thereby creating an expectation for much higher overlap between different sampling techniques. The reality from Figures~\ref{fig7} and~\ref{fig8} contradicts this expectation, at least for intermediate- and low-probability LMs. 

The same comparison of the LVs in the RBM energy function at earlier stages of the training (20 epochs, which corresponded to the 33.3\% classification error) is shown in Figure~\ref{fig9}. As was observed

\vspace{-12pt}

%Fig 9 
 \noindent in Figure~\ref{fig5}b, at these early stages of training, the energy landscape is not that complicated yet, and the number of LVs is much smaller than the one at later stages of training (e.g., 145 LVs after 20 epochs compared to 5059 LVs after 1400 epochs). After 20 epochs, only 20.5\% of LVs (of any energy) found by the D-Wave were missed by Gibbs sampling, and only 9\% of LVs (of any energy) found by Gibbs sampling were missed by the D-Wave (recall Figure~\ref{fig6}a,b). This is confirmed by the much more significant similarities of the dark and light histograms in Figure~\ref{fig9} compared to Figures~\ref{fig7} and \ref{fig8}. However, while the histograms overlap entirely for the most low-energy/high-probability states, the D-Wave and Gibbs sampling still miss a non-negligible percentage of LVs found by the other technique in the intermediate- and high-energy part of the spectrum. 

A qualitatively similar trend was observed for the 20th epoch, using $kG=100$ for the LV search (Figure~\ref{fig10}). At those earlier training stages, many intermediate-to-low probability LMs may correspond to a partially learned TP or spurious LMs, which must be sampled when minimizing the distance between the model and the target probability distributions. As follows from Figures~\ref{fig7}--\ref{fig10}, the ability of the classical Gibbs sampling and the D-Wave QA to sample those states is substantially different, especially at the later stages of the training.
\vspace{-12pt}

%Fig 10 
Finally, computational cost must be mentioned as an important practical consideration. Quantifying the speed of acquiring a sufficiently large number of samples is not a trivial task and was not one of our objectives. However, the following comments can be made. In a comparable RBM/BM training setting, Noè et al. \cite{noe_quantum_2024} reported that reaching a fixed likelihood value took 28.6 s using quantum annealing versus 247.5 s for classical Gibbs sampling ($\sim$8.6$\times$ faster), and estimated that a fully parallelized quantum implementation could reach $\sim822\times$ speedup over a classical GPU baseline for a 16$\times$16 Boltzmann machine. However, they note this advantage may not scale to larger networks, since parallel embedding could require an exponentially growing number of QPU ``clones''. They also report that per-sample D-Wave time was dominated by setup/readout/post-processing overhead ($\sim$600 {$\upmu$}s total) rather than the anneal itself ($\sim$2 {$\upmu$}s), consistent with our own observation that per-call overhead, not anneal time, is the main practical cost driver. We note that queue time for shared D-Wave cloud access is a separate cost not addressed in \cite{noe_quantum_2024} or characterized here, and remains an open practical consideration for adopting this approach at scale.

\section{Conclusions}\label{sec5}
In this work, the energy function of the trained RBM at different training epochs was successfully embedded into the D-Wave Advantage4.1 hardware having a Pegasus lattice. Because of the larger number of qubits and couplings, the QA can be used for higher-dimensionality RBMs compared to our previous work. 

There had been a justified concern that the previously reported ability of the newer generations of the D-Wave hardware to provide a better time to solution \cite{pelofske_comparing_2023} may also cause a reduced ability to obtain a sufficiently high-variance sample from a probability distribution, when many of the D-Wave solution repetitions in a single D-Wave call (out of a maximum of 1000 or 10,000  number of repetitions allowed by the particular version of the hardware) would return repeatedly the GS (or repeated states from a limited number of LVs), thereby failing to sample the needed variety of important states (and LVs). While direct comparison of the Chimera and the Pegasus (D-Wave Advantage4.1) hardware was impossible for our group and was outside the scope of this work, the following relevant question was successfully addressed by this work---can shorter QA times increase the variance of the sample? While the general trend of $N_{\text{LV}}$ increasing with decreasing the annealing time has been qualitatively confirmed, quantitatively, for the RBM and the training dataset investigated in this work, no significant increase in $N_{\text{LV}}$ could have been achieved by changing the annealing time.

When the classical sampling was conducted at conditions like those used in sampling during CD-\textit{k} training, at any training epoch, the states sampled by the D-Wave belonged to a somewhat higher, but comparable, number of LVs relative to those from the Gibbs sampling. This was most pronounced when the classical sampling was done at $kG=1$. However, despite similar numbers of found LVs, many of those LVs turned out to be different for the D-Wave and Gibbs sampling.

The abovementioned difference could inspire an expectation that the two techniques tend to sample not entirely the same parts of the configuration space, thereby offering potentially complementary sampling benefits when used in Contrastive Divergence training of RBMs. Instead, the two techniques were found in this work to be less complementary and more overlapping for high-probability states (i.e., for high-probability LVs). Nevertheless, many of the potentially “important” LVs, those having intermediate (in addition to those having low) probability values at their bottom, were missed by one of the sampling techniques while found by the other.

When comparing the D-Wave to more “aggressive” Gibbs sampling (larger values of the \textit{k}G used), $N_{\text{LV}}$ found by both techniques became even closer. However, the fraction of LVs found by both sampling techniques, relative to the fraction found by only one technique and missed by the other, has not changed substantially. The energy distribution of those LVs that overlap and those that are different between the two techniques was qualitatively (and even quantitatively) very similar for $kG=1$ and larger \textit{k}G values. Each technique misses more of the “important” LVs sampled by the other technique at later rather than earlier training epochs, which is precisely the stage of the training when modest improvements to sampling quality could make meaningful differences for the final RBM trainability. 

The results of this work communicate a reasonable optimism that the D-Wave and the classical Gibbs sampling could complement each other, allowing one to obtain a more representative sample for the given number of sampled states at each training epoch. For example, a hybrid approach could combine Gibbs sampling, which efficiently explores broad local valleys, with quantum annealing, which may contribute additional sampled states from relatively low-energy but narrow local valleys that are underrepresented by classical sampling, thereby improving the overall coverage of the most important parts of the RBM energy landscape. A logical next step in future work is to investigate such hybrid sampling strategies within the contrastive divergence training and evaluate their effect on the classification error, the training convergence rate, or both. 

\vspace{6pt}

\funding{This material is based upon work supported by the National Science Foundation under Grant No. CCF-2211841. Any opinions, findings, conclusions, or recommendations expressed in this material are those of the author(s) and do not necessarily reflect the views of the National Science Foundation.}

\dataavailability{The original contributions presented in this study are included in the article. Further inquiries can be directed to the corresponding author(s).}

\conflictsofinterest{The authors declare no conflicts of interest.} 

\abbreviations{Abbreviations}{%
The following abbreviations are used in this manuscript:\\
\noindent 
\vspace{-10pt}
\setlength{\LTleft}{0pt}
\setlength{\LTright}{\fill}
\begin{longtable}{@{}ll}
RBM & Restricted Boltzmann Machine\\
MCMC&Markov Chain Monte Carlo\\
kG & Number of Gibbs Sampling Steps\\
CD-k & kG-step Contrastive Divergence\\
QC & Quantum Computer\\
QA & Quantum Annealer\\
LV & Local Valley\\
GS & Ground State\\
TP &Training Pattern\\
$N_{\text{LV}}$ & Number of Local Valleys\\
TTS & Time To Solution\\
SF & Scale Factor\\
$N_v$ & Number of Visible Units\\
$N_h$ &  Number of Hidden Units\\
$P_{\text{GS}}$ & Probability of Finding the Ground State\\
$N_{\text{TP}}$ & Number of Training Patterns\\
\end{longtable}
}

\reftitle{References}

\end{document}